\documentclass[onecolumn]{ametsocV6.1}
\nolinenumbers 
\renewcommand{\internallinenumbers}{}

\newlength\savedwidth

\title{Spatiotemporal Forecasting in Climate Data Using EOFs and Machine Learning Models: A Case Study in Chile.}

\authors{ Mauricio Herrera,\aff{a}\correspondingauthor{Mauricio Herrera, mherrera@udd.cl} 
	Francisca Kleisinger ,\aff{a}
	Andrés Wilson,\aff{a}}

\affiliation{\aff{a}{Faculty of Engineering. Universidad del Desarrollo}\\
}

\abstract{
	Effective resource management and environmental planning in regions with high climatic variability, such as Chile, demand advanced predictive tools. The success in these areas heavily relies on accurately interpreting and forecasting climatic patterns. This study addresses these challenges by employing an innovative and computationally efficient hybrid methodology that integrates machine learning (ML) methods for time series forecasting with established statistical techniques. The spatiotemporal data undergo decomposition using time-dependent Empirical Orthogonal Functions (EOFs), denoted as \(\phi_{k}(t)\), and their corresponding spatial coefficients, \(\alpha_{k}(s)\), to reduce dimensionality. Wavelet analysis provides high-resolution time and frequency information from the \(\phi_{k}(t)\) functions, while neural networks forecast these functions within a medium-range horizon \(h\). By utilizing various ML models, particularly a Wavelet–ANN hybrid model, we forecast \(\phi_{k}(t+h)\) up to a time horizon \(h\), and subsequently reconstruct the spatiotemporal data using these extended EOFs.This methodology is applied to a grid of climate data comprising 6355 points covering the entire territory of Chile. It transitions from a high-dimensional multivariate spatiotemporal data forecasting problem (involving 6355 time series) to a low-dimensional univariate time series forecasting problem (requiring only a few dozen forecasts). Additionally, cluster analysis with Dynamic Time Warping for defining similarities between rainfall time series, along with spatial coherence and predictability assessments, has been instrumental in identifying geographic areas where model performance is enhanced. This approach also elucidates the reasons behind poor forecast performance in regions or clusters with low spatial coherence and predictability. By utilizing cluster medoids, the forecasting process becomes more practical and efficient. This compound approach significantly reduces computational complexity while generating forecasts of reasonable accuracy and utility.
}

\begin{document}

\maketitle

%
%
%
\statement
	 The approach outlined in this study facilitates the transition from a high--dimensional multivariate spatiotemporal data forecasting problem to a low-dimensional univariate time series forecasting problem. This transition substantially reduces computational complexity while yielding reasonably accurate forecasts and enhances our ability to interpret and predict climatic patterns across the entire territory and over medium-term temporal horizons, despite its high climatic variability.

%
%

%
\section{Introduction}

Climate change's growing complexity and urgency demand refined yet practical prediction tools, especially in vulnerable regions like Chile with high climatic variability. Effective resource management and environmental planning hinge on our ability to decipher and anticipate climatic patterns, directly impacting agricultural planning and water management. Additionally, understanding precipitation, temperature, and other climatic variables shapes crucial policies concerning climate change and environmental protection. Data-driven analysis guides decision-making towards effective mitigation and adaptation strategies ~\citep{Seneviratne_etal2018,IPCC_2014}.

However, the intricate spatial, temporal, and spatiotemporal correlations in environmental data pose a significant challenge in capturing these dependencies ~\citep{cressie2011statistics}. Understanding and modeling these patterns are crucial for advancing climate research and prediction. Fortunately, a convergence of interests and expertise is emerging. Climate researchers, particularly those in numerical weather prediction, atmospheric physics, extreme events, and climate change, are increasingly turning to Machine Learning (ML) to enhance modeling and prediction ~\citep{atmos13020180}. Similarly, ML researchers recognize the relevance of their work in addressing climate challenges, especially in numerical weather prediction ~\citep{lam2023learning,wong2023deepmind, cao2021spectral}. This collaboration has the potential to unlock ML's capabilities for modeling complex dynamical systems in climate science.

This study employs a hybrid approach for spatiotemporal climate data analysis and forecasting. We harness empirical orthogonal functions (EOFs) for dimensionality reduction, wavelet analysis for high-resolution time-frequency information, and deep neural networks (DNNs) for forecasting. This combined approach capitalizes on the strengths of each technique to tackle the complexities inherent in climate data analysis.

Pioneered in meteorology by ~\cite{lorenz1956empirical}, EOF analysis has become a cornerstone for understanding spatiotemporal climate variability. As explored in seminal works like ~\cite{Preisendorfer_1988}, EOFs reveal orthogonal patterns of variability, each explaining distinct portions of data variance. Notably, the first EOF captures the most significant variance, followed by subsequent ones with diminishing contributions. This inherent efficiency – retaining key information while minimizing complexity – makes EOFs ideal for empirical climate modeling.

The use of Empirical Orthogonal Functions (EOFs) has limitations, as they often lack clear physical significance ~\citep{EmpiricalOrthogonalFunctionsTheMediumistheMessage}. Despite their utility in capturing dominant variance patterns, EOFs may not inherently represent distinct physical processes, making it challenging to distinguish empirical modes from genuine physical phenomena ~\citep{newman1995caveat}. Despite these challenges, EOFs remain valuable for making reasonable predictions with a limited number of data modes. By distilling spatiotemporal variability into a manageable set of orthogonal patterns, EOFs enable efficient empirical modeling and prediction, demonstrating their practical importance in forecasting massive spatiotemporal data. In our study, we utilize EOF analysis alongside ML and wavelets mainly for forecasting, without aiming to distinguish data--driven modes from physically meaningful structures.

This study utilizes a grid comprising 6355 points at a resolution of $0.25\times 0.25$ degrees, covering the entirety of Chile. Each point is associated with a time series spanning from 1980 to 2022, encompassing climatic variables such as daily accumulated precipitation, maximum, mean, and minimum daily temperature, evapotranspiration, etc.

By employing Singular Value Decomposition (SVD), these datasets undergo factorization into Empirical Orthogonal Functions (EOFs) that encapsulate temporal information $\phi_{k}(t)$ (note that we use empirical temporal orthogonal functions, since they are obtained from an empirical temporal covariance matrix) and the corresponding spatial coefficients $\alpha_{k}(s)$, which capture spatial information ~\citep{ hannachi2007empirical}. 

A Wavelet--ANN Hybrid model for forecasting, constructed upon wavelet transform using the \textit{Maximal Overlap Discrete Wavelet Transform} (MODWT) algorithm developed by ~\cite{Anjoy}, facilitates the forecasting of EOFs $\phi_{k}(t + h)$ over a horizon $h$. The spatiotemporal data is then reconstructed utilizing this extended temporal component over a horizon $h$.

This methodology facilitates the transition from a high-dimensional multivariate spatiotemporal data forecasting problem (in this case, entailing forecasting using 6355 time series corresponding to grid points) to a low-dimensional univariate time series forecasting problem (in this case, up to a couple of dozens of forecasts), significantly reducing computational complexity while yielding forecasts of reasonable utility.

To account for the extensive climatic variability of Chile, we use cluster analysis based on time series. We group time series of rainfall with manifest similarities measured by distances based on \textit{Dynamic Time Warping} (DTW).  The structure of clusters or geographic segmentation of similar rainfall patterns is very stable over time.

To make forecast with the proposed methodology, we use the medoids of each cluster, making forecasting in each geographic zone more practical and effective. We conduct these forecasting tests to demonstrate that the proposed methodology has accurately captured the patterns of precipitation behavior over time and space. Additionally spatial coherence and predictability assessments for each clusters, has been instrumental in identifying geographic areas where model performance is enhanced. This approach also elucidates the reasons behind poor forecast performance in regions or clusters with low spatial coherence and predictability.

\section{Materials and methods}
\subsection{Data}

The data were obtained from ERA5 in ~\citep{Era5}, which is part of the Copernicus Climate Change Service (C3S) provided by the European Union and produced by the European Centre for Medium-Range Weather Forecasts (ECMWF). ERA5 offers reanalyzed climatic and meteorological data, covering the period from 1950 to the present, providing detailed information on a wide range of atmospheric, terrestrial, and oceanic variables. It is known for its high spatial resolution and temporal resolution, making it widely used in climate research, environmental studies, and meteorological modeling applications. 

The data used in this study form a grid of 6355 points located between the geographical coordinates - latitude $-17.5°$ to $-56.0°$ and longitude $-76.0°$ to $-66.0°$ with resolution of $0.25 \times 0.25$ degrees. Each point is associated with historical data containing time series of climatic variables such as maximum, mean, and minimum temperature, precipitation, evapotranspiration, etc. Additionally, each point is characterized by its geographical coordinates – longitude and latitude.

Let $X(s,t)=\{x(s_i,t_j\}$ be a data structure for a spatiotemporal variable $x$ (e.g. precipitation, temperature, etc.) with $i=1\ldots n$  and $s_i=($long$_i$, lat$_i$, e$_i)$. Where $i=1\ldots, n$, long$_i$ represents longitude, lat$_i$ represents latitude, and e$_i$ represents elevation of a location $i$ (see Table~\ref{table:1} ). So, $X(s,t)$  consists of $n$ locations, each having an associated time series with $p$ records. In this study, we consider data corresponding to $41$ longitude values and $155$ latitude values, creating a grid of $n=6355$ spatial points. For each of them, there are  $p = 26665$ time values, corresponding to daily records between 1980 and 2022.

\begin{table}[!ht]
	\centering
	\caption{
		{\bf Tidy structure of spatiotemporal data for analysis.}}
	\begin{tabular}{lllll}
      \topline
		& $t_1$         & $t_2$         & $\cdots$ & $t_p$         \\      \botline    
		$s_1$   & $x_{s_1,t_1}$ & $x_{s_1,t_2}$ & $\cdots$ & $x_{s_1,t_p}$ \\
		$s_2$   & $x_{s_2,t_1}$ & $x_{s_2,t_2}$ & $\cdots$ & $x_{s_2,t_p}$ \\
		$\vdots$ & $\vdots$      & $\vdots$      & $\vdots$ & $\vdots$      \\
		$s_n$   & $x_{s_n,t_1}$ & $x_{s_n,t_2}$ & $\cdots$ & $x_{s_n,t_p}$  \\ \botline 
	\end{tabular}
	\label{table:1}
\end{table}

\subsection{Decomposition of spatiotemporal data using EOFs.}
Let $ \mathcal{I}_n$ be a column matrix with $n$ ones. With this matrix, the time mean is expressed as $\bar{x}=\frac{1}{n}X^T\mathcal{I}_n$.

The empirical temporal correlation matrix with dimensions $p\times p$ is written as
\[ C_{p\times p}=\frac{1}{n}X^{T} X - \bar{x}\bar{x}^T =\frac{1}{n} X^THX \]

Where $H=\mathbb{I}-\frac{1}{n}\mathcal{I}_n \mathcal{I}_n^T$ is the so called \textit{centering matrix}.

We transform the data table \ref{table:1} by subtracting the time mean from each value using $Y = X-\mathcal{I}_n\bar{x}^T$. So, $\bar{y} = 0$ and:

$$C_{p\times p}=\frac{1}{n}Y^TY$$

Introducing $Z=\frac{Y}{\sqrt{n}}$ (spatially centered and normalized data) $\Rightarrow C_{p\times p}=Z^TZ$.

Instead of directly calculating the eigenvalues and eigenvectors of the matrix $C_{p\times p}$, the matrix $Z$ is generally (more efficiently) factorized using Singular Value Decomposition (SVD). $Z_{n\times p}=U_{n\times n} D_{n \times p} V_{p \times p} ^T$. Therefore, $Z^T Z = V D^T D  V^T$. Comparing with the spectral decomposition of the matrix $C_{p\times p}=\Phi\Lambda \Phi^T$, where $\Phi$ is the matrix of eigenvectors and $\Lambda$ is a matrix with the eigenvalues of $C_{p \times p}$ on the diagonal, we have $\Phi = V$, $D^T D =\Lambda$, and the Principal Components (PCs) are $\alpha=UD$ ~\citep{jolliffe2002principal}.

The EOFs can be defined as the eigenvectors of covariance matrix $C_{p\times p}$. 
$$C_{p\times p} \phi_k=\lambda_k \phi_k$$

As $C_{p\times p}$ is a nonnegative definite square matrix, the eigenvalues $\lambda_k$ are all nonnegative and the eigenvectors $\phi_k$ form a complete orthonormal basis. So, the centred and normalized data $Z(s,t)=\{z(s_i,t_j\}$ can be represented using a discrete temporal orthonormal basis $\{ \phi_{k}(t_j)\}_{k=1}^{k=K}$ as:

\begin{equation}\label{eq:1}
	z_{{s_{{i}},t_{{j}}}}=\sum _{k=1}^{K}\alpha_{{k}} \left( s_{{i}}
	\right) \phi_{{k}} \left( t_{{j}} \right), i=1,\ldots,n;j=1,\ldots,p
\end{equation}

Where $K=min(n,p)$ and $\alpha_{{k}} \left( s_{{i}}\right)$  is the coefficient corresponding to the $k$ --th basis function $\phi_{{k}} \left( t_{{j}} \right)$ at spatial location $s_i$. It is noteworthy that the scalar coefficient $\alpha_{{k}} \left( s_{{i}}\right)$ depends solely on the location and not on time, whereas the temporal basis function $\phi_{{k}} \left( t_{{j}} \right)$ is independent of space. The rationale behind this decomposition is theoretically grounded in the Karhunen-Loève expansion ~\citep{Preisendorfer_1988}.

The coefficients $\alpha_k$ in this expansion can be calculated using
$$\alpha_k=Z \cdot \phi_k$$
Alternatively, using matrices
$$\alpha = ZV=UD$$ 
So, the expansion coefficients $\alpha_{{k}} \left( s_{{i}}\right)$ are the spatial Principal Components (PCs). 

It follows from the fact that the $\phi_{{k}}$ are orthonormal eigenvectors of $C_{p\times p}$ that the PC are mutually uncorrelated and:

$E(\alpha_{{k}} \left( s_{{i}}\right))=0$

$Var[\alpha_{{1}} \left( s_{{i}}\right)] \geq Var[\alpha_{{2}} \left( s_{{i}}\right)] \geq \cdots \geq Var[\alpha_{{K}} \left( s_{{i}}\right)] \geq 0$

$Cov[\alpha_{{k_1}} \left( s_{{i}}\right), \alpha_{{k_2}} \left( s_{{i}}\right)] = 0$ for all $k_1\neq k_2$,

So, considering the SVD, the functions $\phi_k$ are given by $V_k$, where $V_k$ represents the $k$-th column of the matrix $V$ in the SVD of $Z$. The normalized spatial coefficients are $\alpha_k = Z\cdot V_k= U_k D$. 

The—potentially truncated—EOFs decomposition $(\bar{K}<K)$ returns for each spatial location $s_i$, corresponding to the original observations, $\bar{K}$ random coefficients $\alpha_{{k}}$. These coefficients can be spatially modelled and mapped on a regular grid solving an interpolation/regression task \citep{Amato}.

\subsection{Forecasting EOFs.}

In this study, we forecast the EOFs $\phi_{k}(t)$, in principle, to an arbitrary horizon $h$ using various forecasting techniques. Specifically, the Wavelet-ANN Hybrid Model, as introduced by ~\cite{Anjoy}, consistently yields superior results in forecasting. Leveraging this model, we generate forecasts to predict $\phi_{k}(t+h)$ with a horizon $h$. Subsequently, utilizing this function and the spatial coefficients $\alpha_{k}(s)$, we reconstruct the complete spatiotemporal data. The underlying hypothesis suggests that these spatial coefficients should undergo minimal changes since significant alterations in spatial information are not anticipated within the forecast interval $h$ (see Fig. \ref{fig:spatial} for an example). 

This approximation must consider the uncertainty associated with the temporal forecast of the EOFs. Additionally, when selecting a number $\bar{K}$ of EOFs, we must also incorporate errors associated with the dimensionality reduction.

$$
z_{{s_{{i}},t_{{j}}}} \approx \sum _{k=1}^{\bar{K}}\alpha_{{k}} \left( s_{{i}}
\right) \phi_{{k}} \left( t_{{j}}\right), i=1,\ldots,n; j=1,\ldots, p+h
$$
To reconstruct the original approximated spatiotemporal data, but extended to horizon $h$, $\hat{X}(s,t+h)$, we have:

\begin{equation}\label{recons}
\hat{X}(s,t+h) = \sqrt{n} U_{n\times \bar{K} } D_{\bar{K} \times \bar{K}} V^{T}_{\bar{K}\times p+h}+\mathcal{I}_n\bar{x}^T
\end{equation}

Where $\tilde{D}_{\bar{K} \times \bar{K}}$ is obtained from the matrix $D$ in the SVD decomposition by using only $\bar{K}\leq K$ columns and rows, and $\tilde{V}$ is the matrix $V$, but taking  $\bar{K}$ rows extended with $h$ new elements from the forecast.

 Note that here we use the same $\bar{x}$ under the assumption that incorporating more temporal records does not drastically alter the mean value.

\subsection{Some Considerations for the Application of the Proposed Predictive Model.}

\begin{enumerate}
	\item Potential errors in predicting the EOFs \(\phi_{k}(t+h)\) can propagate and impact the reconstruction of spatiotemporal data. To minimize these propagated errors, it is crucial to achieve accurate predictions within an appropriate horizon $h$ by using the most suitable predictive model.
	
	\item If the relationship between spatial components \(\alpha_{k}(s)\) and temporal components \(\phi_{k}(t)\) changes significantly over the chosen prediction horizon, the assumption of using the same spatial coefficients \(\alpha_{k}(s)\) for both \(\phi_{k}(t)\) and the extended \(\phi_{k}(t+h)\) may be invalid. To capture the spatiotemporal correlations and their stability, we segment the data for the entire territory using clusters. Cluster analysis, along with its assessment of spatial coherence (and thus predictability), aims to identify geographic areas where spatial coherence enhances model performance. It also explains poor prediction results in clusters with low spatial coherence and predictability. For more efficient predictions, we use the medoids of the clusters to apply the model. In clusters with low predictability, results can still be managed by reducing the prediction horizon.
	
	\item If the original data contains nonlinearities, EOFs decomposition may not be suitable. It is advised to verify, prior to prediction, the alignment between the real data and the data reconstructed from previously decomposed EOFs. If the alignment is good, the model can be applied to achieve accurate forecasts.
	
	\item The proposed model’s predictions are valid for short to medium horizons. Using historical data with extensive records may include complex nonlinear patterns and significant variations in spatiotemporal conditions that will affect the forecast with this method. For studying historical variations and long-term changes, it is recommended to use models other than the one proposed here (for example, Non-Homogeneous Hidden Markov Models).
\end{enumerate}

\subsection{Analyzing Precipitation Pattern Similarities Using Dynamic Time Warping on Time Series.}

In this section, we compare precipitation time series from 460 strategically selected geographic locations from the data grid and across Chile in search of similarity patterns. These locations are positioned around actual meteorological stations nationwide \citep{cr2_precipitacion}.

The objective of time series comparison methods is to produce a distance metric between two input time series. The similarity or dissimilarity of two time series is typically calculated by converting the data into vectors and calculating the Euclidean distance between those points in vector space. Traditional time series Euclidean Matching is extremely restrictive. However, \textit{Dynamic Time Warping} (DTW) \citep{giorgino2009computing} allows the two curves to match up evenly even though the X-axes (i.e., time) are not necessarily in sync. The rationale behind DTW is to stretch or compress two time series locally in order to make one resemble the other as much as possible. The distance between the two is then computed, after stretching, by summing the distances of individual aligned elements.

\begin{figure}[!ht]
	\centering
	\includegraphics[width=1.0\textwidth]{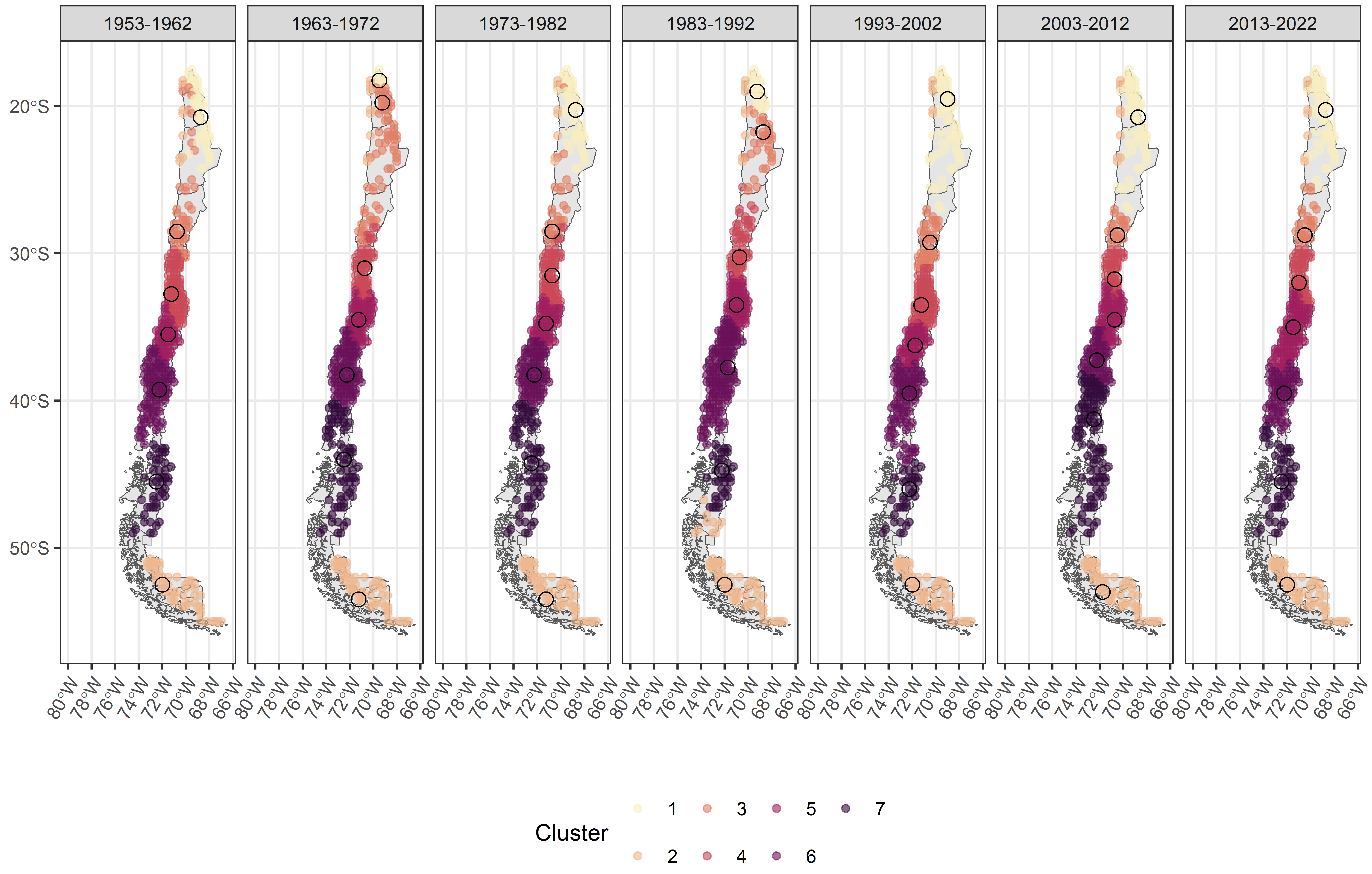}
	\caption[scheme]{ Solution of Seven Clusters Using Hierarchical Method and DTW -- Based Distance. Records are segmented into ten-year intervals to illustrate stability versus slight variations in cluster structure.}
	\label{fig:cluster1}
\end{figure}

We are employing the Dynamic Time Warping (DTW) method for a cluster analysis with two primary objectives:
\begin{enumerate}
	\item To demonstrate that the relationship between precipitation patterns and geographical location is generally stable (stability of spatiotemporal precipitation patterns).
	\item For cluster-based forecasting. By focusing on clusters, we can improve forecast accuracy for the region, for example, by using the representative (medoid) of each cluster for model predictions.
\end{enumerate}

Fig. \ref{fig:cluster1} presents a seven-cluster structure obtained using the hierarchical algorithm with Ward method, and with DTW-based distance. To demonstrate the stability of spatiotemporal patterns, precipitation data from 460 locations across the territory, with time series records from 1953 to 2022, are segmented into 10 -- year intervals, and clusters were constructed for each interval. Only records from the months of May, June, July, and August (MJJA), which correspond to the rainy season in Chile, were considered. This segmentation aims to capture the stability of rainfall patterns and observe any minor changes in the structure of these precipitation patterns.

\begin{figure}[!ht]
	\centering
	\includegraphics[width=1.0\textwidth]{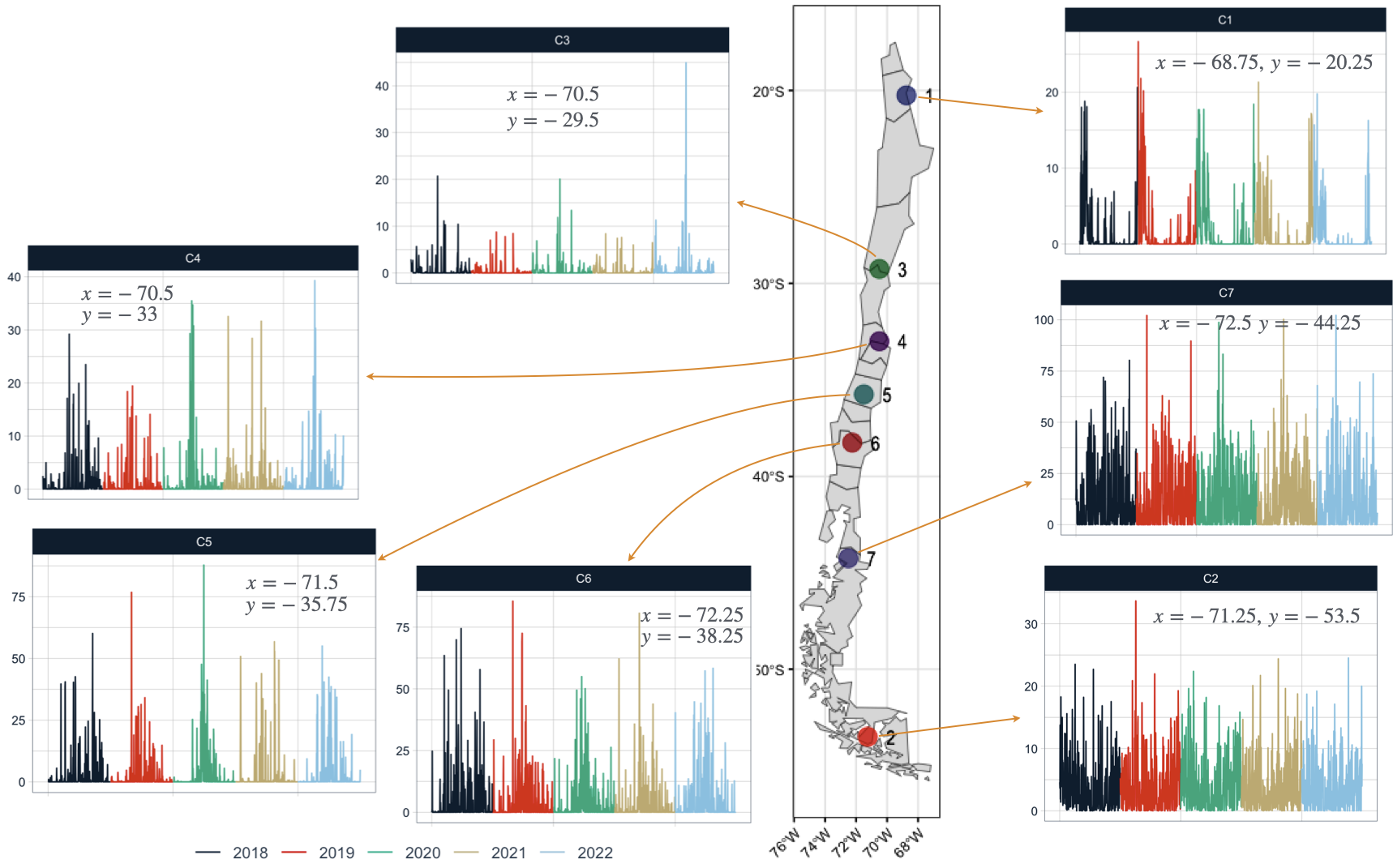}
	\caption[scheme]{Cluster--Specific Precipitation Patterns from 2018 to 2022}
	\label{fig:patterns}
\end{figure}

Selecting a structure with seven clusters ensures a segmentation of the data that provides enough records in each cluster to proceed with EOF analysis and further training of the ML models for time series. This structure effectively captures the specific patterns present in the precipitation data, enabling a detailed and meaningful analysis.

Fig. \ref{fig:patterns} depicts the characteristic precipitation patterns of each cluster. The time series shown in the figure represent the precipitation records from 2018 to 2022 for the medoids, or typical representatives, of each cluster.

\subsection{Spatial coherence and potential predictability.}

The spatial coherence provides a measure of potential predictability at the location scale \citep{moron2007spatial}. Two scores are frequently used to provide empirical estimates of the spatial coherence of seasonal anomalies between locations: $var(SAI)$  - the interannual variance of the standardized anomaly index \citep{katz1986anatomy}, and \textit{DOF} - the number of spatial degrees of freedom \citep{bretherton1999effective}. 

The \textit{SAI} is computed by standardizing the interannual time series at each location (subtracting the mean and dividing by the STD) and then averaging the standardized anomalies spatially across the locations to form an index; it thus gives each location equal weight in the index. The amplitude of the \textit{SAI} for a particular year depends on the size of the correlations between locations, and thus its variance gives a measure of spatial coherence of the field.

\[ var(SAI_i)=var\big[{\frac {1}{n}\sum _{j=1}^{n}{\frac {x_{{i j}}-\bar{x}_{{j}}}{\sigma_{{j}}}}}\big]
\]

where $x_j$ is the long--term time mean over $i=1\ldots l$ years and $\sigma_j$ is the interannual standard deviation for location $j$. The $var(SAI)$ is a maximum when all locations are perfectly correlated, $var(SAI) =1$, and a minimum when the locations are uncorrelated, resulting in a $var(SAI)=\frac{1}{n}$.

The DOF gives an empirical estimate of the spatial coherence in terms of empirical (spatial) orthogonal functions, with higher values denoting lower spatial coherence:

\[ {\it DOF}={\frac {{n}^{2}}{\sum _{j=1}^{n}{\lambda_{{j}}}^{2}}}
 \]
where $\lambda_j$ are the eigenvalues of the correlation matrix formed from the location seasonal--mean time series and $n$ is the number of locations.

\begin{table}[!ht]
	\centering
	\caption{
		{\bf Table of Clusters for Degrees of Freedom (DOF) and Variance of the Standardized Anomaly Index (var(SAI)) for Accumulated Rainfall, Rainfall Intensity, and Rainfall Frequency during the MJJA period for the years 1980--2022}}
\begin{tabular}{llllllll} 
	\topline
	~ & ~ & DOF & ~ & ~ & var(SAI) & ~ & ~ \\ \hline
	~ & N° & RAm & RI & RF & RAm & RI & RF \\ \hline
	Clusters & ~ & ~ & ~ & ~ & ~ & ~ & ~ \\ \hline
	1 & 64 & 2,70 & 5,94 & 2,70 & 0,52 & 0,29 & 0,54 \\ \hline
	2 & 64 & 2,58 & 5,57 & 2,47 & 0,51 & 0,26 & 0,53 \\ \hline
	3 & 48 & 1,39 & 2,58 & 1,81 & 0,84 & 0,59 & 0,73 \\ \hline
	4 & 75 & 1,20 & 1,65 & 1,46 & 0,91 & 0,77 & 0,82 \\ \hline
	5 & 53 & 1,16 & 1,51 & 1,21 & 0,93 & 0,81 & 0,91 \\ \hline
	6 & 80 & 1,25 & 1,81 & 1,34 & 0,89 & 0,73 & 0,86 \\ \hline
	7 & 76 & 2,05 & 2,97 & 1,92 & 0,64 & 0,49 & 0,69 \\ \hline
	Altitude class & ~ & ~ & ~ & ~ & ~ & ~ & ~ \\ \hline
	0-500 m & 293 & 3,46 & 5,78 & 3,68 & 0,27 & 0,21 & 0,28 \\ \hline
	500-1500 m & 87 & 2,56 & 3,66 & 2,90 & 0,50 & 0,39 & 0,49 \\ \hline
	$>$1500 m & 80 & 3,41 & 6,91 & 3,53 & 0,42 & 0,26 & 0,37 \\ \hline
	All locations & 460 & 4,11 & 6,68 & 4,56 & 0,24 & 0,20 & 0,22 \\ \hline
\end{tabular}
	\begin{flushleft}  
	\end{flushleft}
	\label{table:varsaidof}
\end{table}

Table \ref{table:varsaidof} displays the \textit{DOF} and $var(SAI))$ for Accumulated Rainfall (\textbf{RAm}), Rainfall Intensity (\textbf{RI}: calculated as the total millimeters of rain divided by the number of rainy days. Where daily precipitation exceeds 1 mm), and Rainfall Frequency (\textbf{RF}: calculated as the number of rainy days divided by the total number of days in the period) from 1980 to 2022 (considering only MJJA) across 460 locations. Additionally, it considers the cluster structure based on DTW.

\begin{figure}[!ht]
	\centering
	\includegraphics[width=0.5\textwidth]{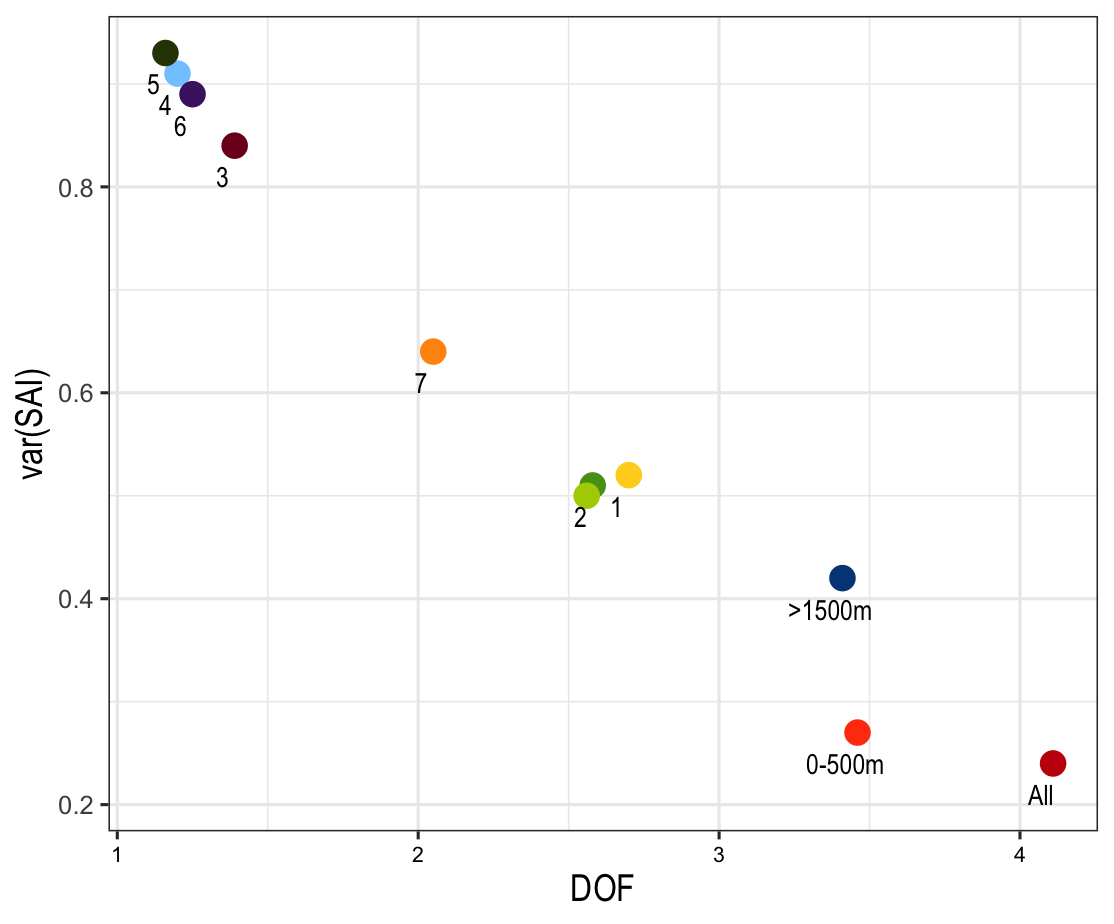}
	\caption[scheme]{DOF and $var(SAI)$ metrics for Rainfall amount (RAm) by clusters (1 - 7). Groups ``All'', ``0 -- 500m'', ``500 -- 1500m'' and ``$>500$m'' are included.}
	\label{fig:DOF-VARSAI}
\end{figure}

Fig. \ref{fig:DOF-VARSAI} shows the \textit{DOF} and $var(SAI)$ metrics for RAm.  The cases ``All'' (all locations), ``0 -- 500m'' (locations with elevations between 0 -- 500m), and ``$>500$m'' (locations with elevations above 1500 m) have the highest DOF and the lowest var(SAI) values. This indicates that the spatial coherence of these locations is lower than in other clusters. For ``All'' and ``0-500m'', this can be attributed to the large number of locations considered in the calculation (460 and 293, respectively), resulting in a diversity of microclimates. For ``$>500$m'', the lower coherence is explained by the effect of orographic rainfall at these elevations.

On the other hand, clusters 3, 4, 5, and 6 exhibit values indicating better spatial coherence. The DOF for these clusters is close to the minimum (\textit{DOF} = 1) and $var(SAI)$ is close to the maximum ($var(SAI)=1$). These clusters correspond to the central and central--northern regions of Chile. These values indicate that locations within these clusters have very similar behavior patterns. Given the high spatial coherence in these clusters, predictive models are likely to capture their precipitation patterns accurately. However, this is not the case for clusters 1, 2, and possibly 7, which have \textit{DOF} and $var(SAI)$ values indicating low spatial coherence. It is worth noting that the clustering structure found generates groups with significantly lower \textit{DOF} and higher $var(SAI)$ compared to segmentations based on elevation ranges.

Considering RI results in a general decline in both indicators' values across all clusters, with more significant effects in some. Clusters 1 and 2, which already had the lowest coherence for accumulated precipitation, show similar trends in rainfall intensity, aligning more with``All'', ``0 -- 500m'', and ``$>500$m''. This indicates greater variability in rainfall intensity within these clusters compared to average precipitation, with worsening effects especially pronounced in clusters 1 and 2. For clusters with higher coherence, cluster 3 deteriorates and diverges from previously better--performing clusters (4, 5, and 6). This is due to its northern location, adjacent to cluster 1, both situated in the arid regions of North Grande and North Chico in Chile, which experience scarce and infrequent rainfall.

For RF, clusters 1 and 2 continue to exhibit the worst coherence, while clusters 4, 5, and 6 maintain the best.

In Summary:
\begin{itemize}
	\item Clustering results in groups with lower \textit{DOF} and higher $var(SAI)$ compared to separation by altitude ranges, consistent with capturing micro-climates unique to the clusters.
	\item Clusters 4, 5, and 6 show the best spatial coherence. 
	\item The low coherence in rainfall intensity for cluster 1 is due to its location in northern Chile, an area with very scarce precipitation.
	\item High coherence in clusters 4, 5, and 6 suggests that predictive models can accurately capture their patterns and perform well.
	\item Low to medium spatial coherence in clusters 1, 2, 3, and 7 implies challenges in using predictive models effectively.
\end{itemize}

\section{Results}

\subsection{Forecasting Precipitation Across the Chile Territory.}

To test the forecast using the described method, we will consider precipitation records from a five-year period between 2018 and 2022 for all data grid points covering the country. The first four years (2018–2021) are used as training data to fit the predictive model, and the forecast is made for the entire year 2022, representing a horizon of \(h = 365\) days.

Both training and forecasting are performed independently for all grid points within each cluster's defined geographic area. Recall that the clusters are constructed from a sample of 460 points (out of a total of 6355 grid points) near the precipitation measurement stations. The cluster structure segments this data sample by capturing spatiotemporal correlations. Thus, similar precipitation patterns in the records (similarity defined using DTW distances) are associated with specific geographic areas. Moreover, this cluster structure is stable enough for analysis. Here, ``stable enough'' means a minimal variability of the cluster structure during the period chosen for model fitting and the forecast horizon considered. These stable spatiotemporal correlations allow the use of extended EOFs for forecasting to reconstruct the data. We thus consider the forecast in seven geographic regions, including all grid points falling within each cluster-defined region, to fit the predictive model.

Depending on the cluster, the number of data grid points (time series) considered varies, as each cluster has a different number of locations within its delineated geographic zone. Additionally, the number of EOFs to be considered also depends on the chosen cluster. The criterion is that this number of EOFs should ensure more than 80\% of the explained variance. Clusters with better DOF and Var(SAI) indicators require fewer EOFs for the analysis compared to clusters with poorer indices. Thus, clusters labeled 3 to 6, which present better metrics, require between 3 to 5 EOFs to explain more than 80\% of the variance, while clusters 1, 2, and 7 require more than 5 EOFs to achieve this.

We decomposed the data using Singular Value Decomposition (SVD), obtaining Empirical Orthogonal Functions (EOFs) and Principal Components (PCs). In practice, this is akin to decomposing the data into a series using EOFs as basis functions, where the coefficients represent the PCs. We selected \(\bar{K} = 5 \)  to \(\bar{K} = 7 \) PCs for each cluster-defined geographic region, accounting for more than 80\% of the explained variance in each case, and used their corresponding EOFs to forecast the spatiotemporal data for each region.

\begin{figure}[!ht]
	\centering
	\includegraphics[width=1.0\textwidth]{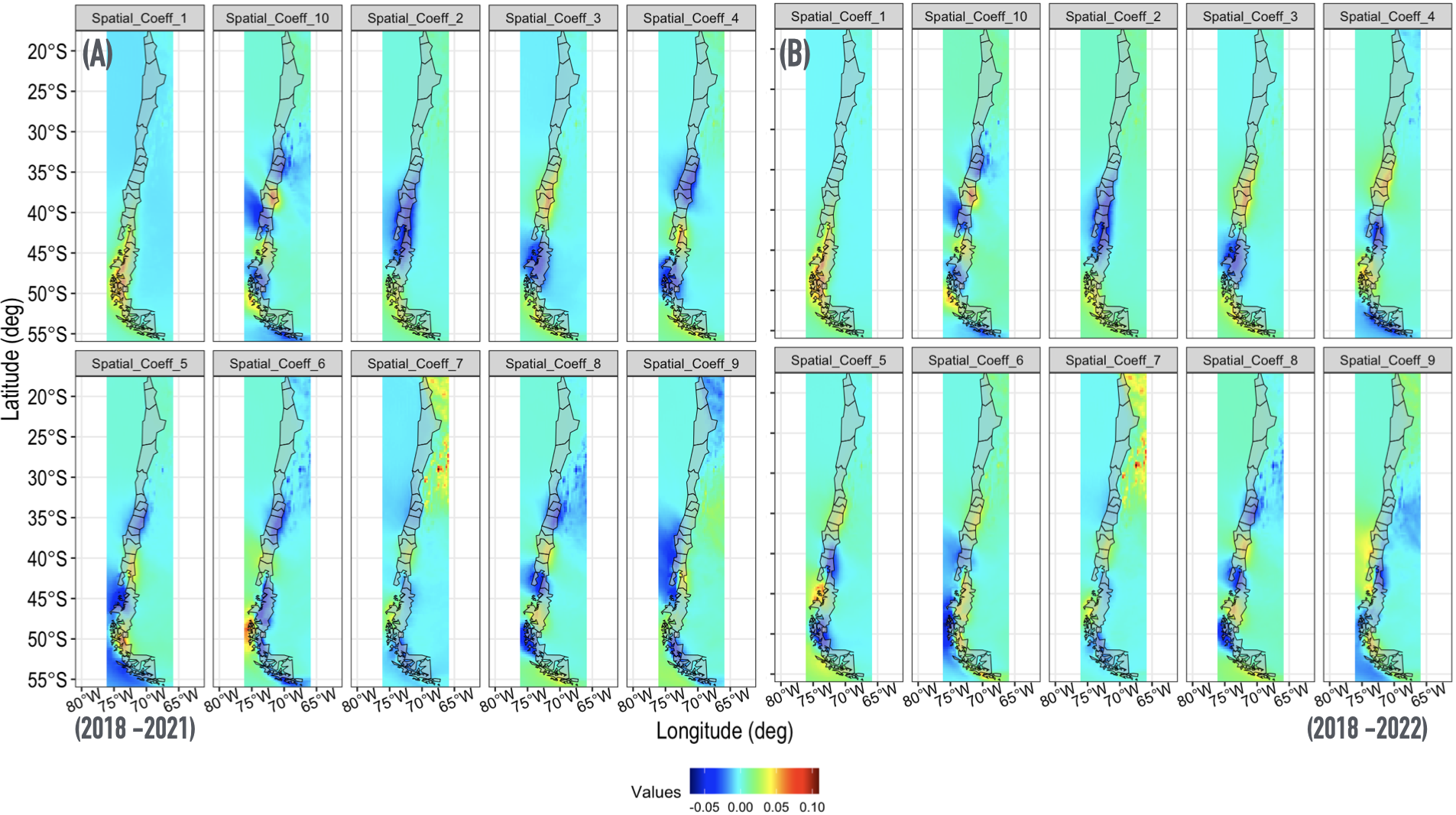}
	\caption[scheme]{ Comparison of the top 10 spatial coefficients from EOF decomposition of precipitation data for 2018–2021 (panel A) and data including 2022 (panel B).}
	\label{fig:spatial}
\end{figure}

The coefficients $\alpha_{k}(s)$ (PCs), which solely depend on spatial coordinates, demonstrate minimal variability when extending the forecasting horizon to a year (i.e.,to 2022), as supported by empirical evidence. Upon meticulous examination of EOFs and their associated spatial coefficients across various temporal intervals in spatio--temporal data decomposition, it becomes evident that the first spatial coefficients (i.e., the first PCs) exhibit negligible variation across these different studied time intervals. Fig. \ref{fig:spatial} illustrates this by comparing the first 10 spatial coefficients calculated for the dataset between 2018 and 2021 with those calculated when incorporating the year 2022.
 
By leveraging decomposition by EOFs, the spatiotemporal forecasting problem transitions into a univariate time series forecasting problem, as EOFs are inherently uncorrelated time series. This enables the utilization of various well--developed methodologies for time series forecasting. As a baseline, we employed classical autoregressive models available in the R package ``modeltime'' \citep{Dancho} (see Table 	\ref{table:Autoregressive}). Additionally, we utilized Deep Learning autoregressive models, including DEEPAR, which is a DL architecture based on a Long Short--Term Memory (LSTM) Recurrent Neural Network ~\citep{Flunkert_etal2017}, DEEP STATE, an approach to probabilistic time series forecasting that combines state space models with deep learning ~\citep{NEURIPS2018_5cf68969}, NBEATS, a deep neural architecture based on backward and forward residual links and a very deep stack of fully--connected layers ~\citep{oreshkin2019nbeats}, and Gaussian Process (GP) Forecast, a DL architecture that automatically selects the optimal kernel in Gaussian process analysis of time series, while also providing reliable estimation of the hyperparameters ~\citep{NEURIPS2019_0b105cf1}, for forecasting purposes.

\begin{table}[!ht]
	\centering
	\caption{
		{\bf Accuracy table for some autoregressive models.}}
	\begin{tabular}{lllllll}
		\topline
		& Model            & MAE     & MAPE     & MASE	& SMAPE   & RMSE        \\  \botline    
		$1$ & ARIMA        & $0.01$ & $785.61$ & $1.04$ & $77.03$ & $0.02$       \\
		$2$ & PROPHET      & $0.01$ & $806.55$ & $1.05$ & $78.73$ & $0.02$        \\
		$3$ & GLMNET       & $0.01$ & $811.93$ & $1.09$ & $80.53$ & $0.02$         \\
		$4$ & SVM--RBF      & $0.01$ & $753.24$ & $1.05$ & $79.32$ & $0.02$         \\
		$5$ & BOOST--TREE (H2O) & $0.01$ & $941.93$ & $1.12$ & $79.37$ & $0.02$         \\
		$6$ & PROPHET--XGBOOST & $0.02$ & $918.38$ & $1.18$ & $83.03$ & $0.02$         \\
		$7$ & RANDOM--FOREST & $0.01$ & $825.46$ & $1.05$ & $77.30$ & $0.02$          \\ \botline 
	\end{tabular}
	\label{table:Autoregressive}
\end{table}

While classic autoregressive models exhibited poor performance, DL models produced more accurate results. Table \ref{table:2} summarizes the key metrics for comparing the models \textbf{(MAE}: Mean absolute error, \textbf{MAPE}: Mean absolute percentage error, \textbf{MASE}: Mean absolute scaled error,\textbf{ SMAPE}: Symmetric mean absolute percentage error, and \textbf{RMSE}: Root mean squared error).

\begin{table}[!ht]
	\centering
	\caption{
		{\bf Accuracy table comparing different Deep Learning models in forecasting the EOF $\phi_1$.}}
	\begin{tabular}{lllllll}
		\topline
		& Model            & MAE     & MAPE     & MASE	& SMAPE   & RMSE        \\  \botline    
		$1$ & WAVELET--ANN [\cite{Anjoy}] & $0.015$ & $2.942$ & $2.349$ & $0.019$ & $0.971$       \\
		$2$ & DEEPAR  [\cite{SALINAS20201181}]    & $0.016$ & $833.$ & $1.25$ & $89.1$ & $0.021$        \\
		$3$ & DEEP STATE [\cite{NEURIPS2018_5cf68969}]  & $0.014$ & $661.$ & $1.08$ & $83.0$ & $0.018$         \\
		$4$ & NBEATS   [\cite{oreshkin2019nbeats}]   & $0.014$ & $813.$ & $1.09$ & $81.3$ & $0.017$         \\
		$5$ & GAUSSIAN PROCESS FORECAST [\cite{NEURIPS2019_0b105cf1}] & $0.014$ & $777.$ & $1.04$ & $77.7$ & $0.017$   \\ \botline 
	\end{tabular}
	\label{table:2}
\end{table}

\begin{figure}[!ht]
	\centering
	\includegraphics[width=1.0\textwidth]{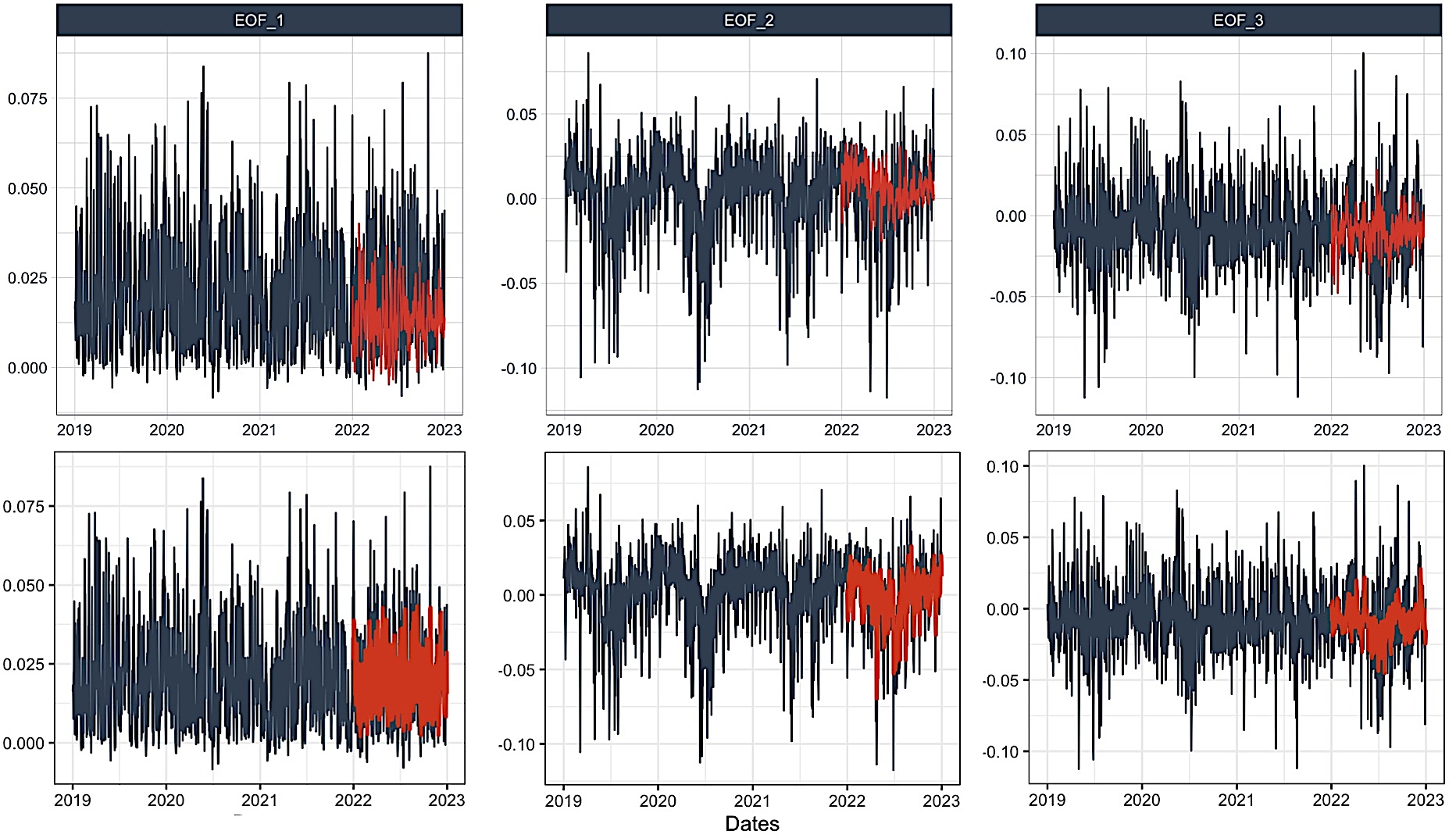}
	\caption[scheme]{The first three EOF functions (in black) and their predictions (in red). The top three panels display predictions of the EOFs using a LSTM--Recurrent Neural Network with ten hidden layers, while the bottom three panels depict predictions using the Wavelet--ANN Hybrid Model.}
	\label{fig:1}
\end{figure}

Fig. \ref{fig:1} depicts a comparison of forecasts for 2022 generated by the Wavelet--ANN Hybrid Model and the DEEPAR model using the first three EOFs. Precipitation records from three years between 2019 and 2021 were utilized as training data for model training.

In the remainder of this article, we will continue using the Wavelet–ANN Hybrid model for forecasting, which is built upon the wavelet transform using the Maximal Overlap Discrete Wavelet Transform (MODWT) algorithm developed by Anjoy et al. ~\cite{Anjoy}. 

For EOFs forescast with the Wavelet--ANN Hybrid model, we employed a Haar filter with 10 wavelet levels (the level of wavelet decomposition), and the size of the hidden layer = 40. Next, we reconstructed the spatiotemporal data for all locations within each cluster.
 
\begin{figure}[!ht]
	\centering
	\includegraphics[width=1.0\textwidth]{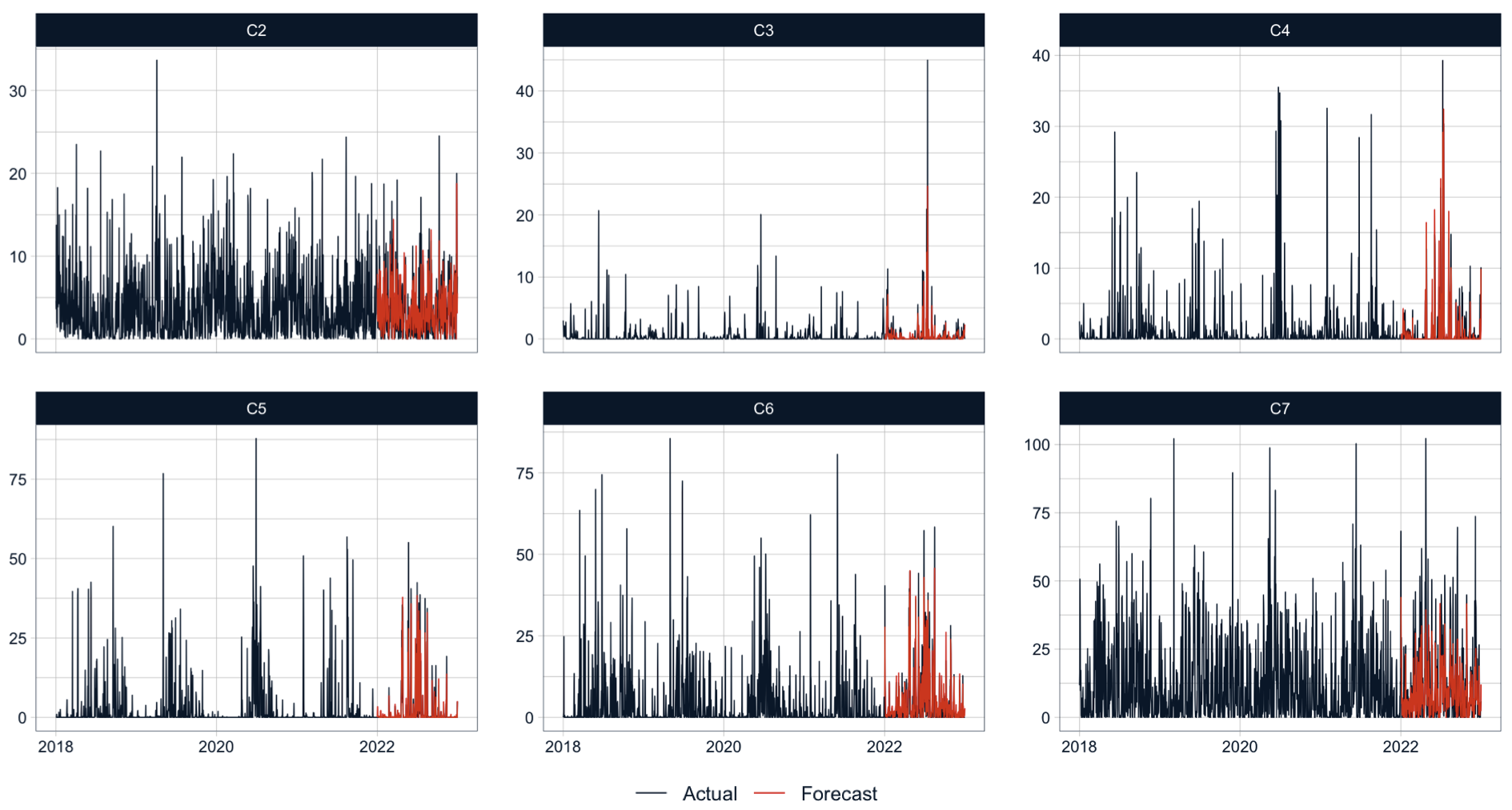}
	\caption[scheme]{The precipitation time series for the medoids of clusters C2 to C7 for 2022 are shown. The black curves represent the actual precipitation records, while the red curves display the forecasts made using the EOF–Wavelet–ANN hybrid model.}
	\label{fig:3}
\end{figure}

Fig. \ref{fig:3} offers a detailed illustration of the 2022 forecast using this method. By conducting the forecast separately for each cluster, we ensure similar precipitation patterns and more stable spatiotemporal relationships.

We primarily use the medoid or representative of each cluster to validate the forecast. The medoid captures a characteristic precipitation pattern for each cluster and, consequently, for the geographic area it represents. The other points within this geographic area are expected to exhibit similar precipitation patterns. However, in larger clusters or those with poorer spatial coherence metrics, some time series within the cluster may deviate from the medoid’s pattern. In these instances, we conduct multiple forecasts at various points within each cluster to thoroughly assess the model's performance.

Table 	\ref{table:3} shows the forecast performance for each cluster medoid using the decomposition of spatiotemporal data into EOFs and capturing temporal patterns with Wavelet-ANN. The table shows the number of grid points (out of a total of 6355 points) within each cluster that were used to train the model. 

To illustrate the complexity of applying the described model, the computation time for each data set corresponding to each cluster is indicated, using a MacBook Pro with a 2.3 GHz Intel Core i9 Eight-Core processor and 16 GB of 2667 MHz DDR4 memory. Predictions can be made for any grid point within the clusters, but for visualizing the predictions, the medoid of each cluster is used.

\begin{table}[!ht]
	\centering
	\caption{
		{\bf Model Performance on the Precipitation Data for the Medoids of the Clusters.}}
		\begin{tabular}{llllllllll}
		\hline
		Cluster &Grid points & MAE & MASE & RMSE & Medoid(Long,Lat)  & Expl.Variance(5EOFs) & Comput.Time & ~ & ~ \\ \hline
		1 &1232 & 1.423551 & 1.615929 & 2.122508 & (-68.75,-20.25) & 0.8148431(7EOFs) & 39.18936 mins & ~ & ~ \\ 
		2 &1542 & 1.609527 & 0.4985094 & 2.280558 & (-71.25,-53.5) & 0.8998416 & 26.27438 mins & ~ & ~ \\ 
		3 &533 & 0.4189356 & 0.444972 & 1.348227 & (-70.5,-29.25) & 0.8478964 & 27.81493 mins & ~ & ~ \\ 
		4 & 448& 0.6333704 & 0.4218802 & 1.516656 & (70.5,-33) & 0.8785737 & 28.12317 mins & ~ & ~ \\ 
		5 &355 & 0.7353927 & 0.2167723 & 2.263974 & (-71.5,-35.75) & 0.8933317 & 27.90131 mins & ~ & ~ \\ 
		6 &580 & 2.022632 & 0.3941201 & 3.10687 & (-72.25,-38.25) & 0.8761673 & 27.94201 mins & ~ & ~ \\ 
		7 &1661 & 6.471836 & 0.6291881 & 10.05184 & (-72.5,-44.25) & 0.8931514 & 28.08024 mins & ~ & ~ \\ 
		~ & & ~ & ~ & ~ & ~ & ~ & ~ & ~ & ~ \\ \hline
	\end{tabular}
	\begin{flushleft}  
	\end{flushleft}
	\label{table:3}
\end{table}

Fig. \ref{fig:3} shows the precipitation time series for the medoids of clusters C2 to C7 for 2022. The locations of the medoids are indicated in table \ref{table:3}. The black curves represent the actual precipitation records, while the red curves show the forecasts made using the previously described method. 

It can be observed that the forecasts appear reasonably accurate for these locations, with more precise results for the medoids of clusters C3 to C6. For the medoids of clusters C2 and C7, the forecasts seem less accurate, consistent with the indicators shown in Figure \ref{fig:3}. 
 
 Time series for cluster C1 are not shown, but the results are less accurate. This is due to the low predictability indicated by the poor \textit{DOF} and $var(SAI)$ metrics for these geographic areas.

\section{Conclusions and Discussion.}

In this study, we presented an innovative approach to analyzing and forecasting spatiotemporal climatic patterns by integrating advanced statistical techniques with machine learning methods. By employing Empirical Orthogonal Functions (EOFs) for dimensionality reduction, we efficiently capture the essential temporal and spatial information within the climate data. The application of wavelet analysis provides high-resolution time-frequency information, enhancing the detail and accuracy of the forecasts. Machine learning methods are leveraged for their powerful predictive capabilities, allowing for the robust forecasting of a truncated number of EOFs over a medium-range horizon. Specifically the Wavelet-ANN Hybrid model, utilizing the Maximal Overlap Discrete Wavelet Transform (MODWT) algorithm, has proven to be effective in forecasting the selected EOFs, thereby enabling the reconstruction of spatiotemporal data with extended temporal components. 

This methodology not only addresses the complexities inherent in climate data analysis but also facilitates the transition from a high-dimensional multivariate forecasting problem to a more manageable low-dimensional univariate forecasting problem. This significant reduction in computational complexity is achieved without compromising the utility and accuracy of the forecasts.

Moreover, the use of cluster analysis with Dynamic Time Warping (DTW) for defining similarities between rainfall time series, along with spatial coherence and predictability assessments, has been instrumental in identifying geographic areas where model performance is enhanced. This approach also provides insights into the reasons behind poor forecast performance in regions or clusters with low spatial coherence and predictability. By using the medoids of the clusters, the forecasting process becomes more practical and efficient.

Overall, this study underscores the importance of combining statistical and machine learning techniques to tackle the intricate challenges posed by climate data analysis. The findings highlight the potential of this hybrid methodology to improve resource management and environmental planning, particularly in regions characterized by high climatic variability like Chile. The robust framework established in this research offers a pathway for more accurate and reliable climatic forecasts, ultimately contributing to better-informed decision-making processes in the face of climate change.

The primary motivation behind this study is to use pragmatic, computationally efficient models that provide useful forecasts while accounting for the climatic variability of an extensive and complex region such as Chile. Given this significant variability, we posit that a single predictive model or tool adaptable to all the intricate details scattered throughout the region is inherently unreliable.

To address the complexities of this problem, we employ a hybrid approach that combines several tools: (1) unsupervised classification (cluster analysis) to capture spatiotemporal correlations using Dynamic Time Warping (DTW)-based distances, (2) Empirical Orthogonal Function (EOF) decomposition on data classified (segmented) by clusters to reduce the problem's dimensionality, and (3) the application of machine learning methods on time series to capture temporal patterns in each geographic zone delineated by the clusters. 

This approach yields localized forecasts for geographic regions exhibiting similar precipitation behavior patterns.

It is important to note that this method can also be applied to the entire dataset for a global forecast of the territory (i.e., without segmenting into geographic zones delineated by clusters) by increasing the number of EOFs (20 or more, to achieve more than 80\% of the explained variance). However, this would result in reliable forecasts in areas with better spatial coherence and predictability, but less reliable predictions in other zones where the data is not well represented. Segmenting the data into similarity patterns is a preliminary step to enhance the local nature of the forecast and can be beneficial for each specified geographic zone.

Furthermore, this approach can be combined with the innovative methodology recently proposed in ~\cite{Amato}, where spatial coefficients can be extended not only across the entire grid but also at any point within the territory using Deep Learning-based regression (interpolation). A dense Feed--Forward Neural Network (FFNN) captures local spatial patterns, facilitating the reconstruction of $\alpha(s)$ at any point. By passing the extended $\phi(t + h)$ over a horizon $h$ to the final layer (recombination layer) of this FFNN for estimating spatial coefficients, the spatiotemporal fields are not only reconstructed at each grid point (potentially at any point, not limited to the grid) but also extended in time through the forecast of the associated time series.

\newpage

\bibliographystyle{ametsocV6}
\bibliography{references}

\newpage








%



\end{document}